%% file: root.tex
\documentclass[letterpaper, 10 pt, conference]{ieeeconf}  

\IEEEoverridecommandlockouts                              

\usepackage{color}
\usepackage{amsmath} 
\usepackage{amssymb}  
\usepackage{mathtools}
\usepackage{booktabs, multirow}
\usepackage{algorithm}
\usepackage{algpseudocode}
\usepackage{comment}

\usepackage{amsthm}
\usepackage{url}
\newtheorem{theorem}{Theorem}  \newtheorem{proposition}{Proposition}
  \theoremstyle{remark}

\title{\LARGE \bf
Physics-Constrained Neural ODEs for Learning Traffic Dynamics and Trajectories from Road Entry/Exit Times*
}
\author{Dennis Bogatov Wilkman$^{1}$, Karl Henrik Johansson$^{1}$, and Matthieu Barreau$^{1}$
\thanks{*This work is supported by the Wallenberg AI, Autonomous Systems and Software Program (WASP) funded by the Knut and Alice Wallenberg Foundation.}%
\thanks{$^{1}$Authors are with the Department of Decision and Control Systems, School of Electrical Engineering and Computer Science, Digital Futures, KTH Royal Institute of Technology, 100-44, Sweden {\tt\small \{dwilkman, kalle, barreau\}@kth.se}}}%

\begin{document}

\maketitle
\thispagestyle{empty}
\pagestyle{empty}


\begin{abstract}
We consider the problem of learning microscopic traffic dynamics and reconstructing vehicle trajectories from section entry/exit timestamps (EET) recorded by cameras and from the positions of a probe vehicle leading the platoon. The dynamics are difficult to recover as each vehicle's position is measured only at two points, and its trajectory between them is unknown. By combining Neural ODEs with a local density approximation, we learn a follow-the-leader model whose speed--density relation is a neural network. With such sparse data, an unconstrained network may learn physically unrealistic relations, such as negative speeds or non-zero speeds on a fully congested road. We therefore introduce a normalization that, applied to a monotone neural network with a learned regime change, enforces the endpoint conditions, guaranteeing an admissible speed--density relation while provably retaining the ability to approximate any admissible one. We validate the method on first- and second-order simulations, recovering the speed--density relation of the first-order model and reconstructing the trajectories of both models, with mean maximum position errors of 2.50~m and 2.51~m, respectively, on a 125~m road segment. On real-world ring-road data, the proposed model still yields an admissible relation, whereas both baselines violate the constraints; the large spread in the data, however, highlights the limitations of a first-order model with homogeneous driver behaviors.
\end{abstract}


\input{Sections/Introduction}
\input{Sections/PreliminariesProblemStatement}
\input{Sections/Method}
\input{Sections/ResultsDiscussion}

\bibliographystyle{ieeetr}
\bibliography{Sections/references}


\end{document}

%% file: Sections/Introduction.tex
\section{Introduction}
Traffic congestion increases emissions and energy consumption and wastes travel time. Most of these losses occur in the congested regime, where traffic slows down, and perturbations are amplified~\cite{ferrara2018freeway}. Control strategies, such as the use of automated vehicles, can dampen these perturbations~\cite{stern2018dissipation,lee2025traffic}, but they require accurate estimates of the traffic state and its dynamics. Recent technological developments, such as probe vehicles and new infrastructure sensors, can be leveraged for this purpose.

From a methodological perspective, traffic dynamics and trajectories have been estimated through fundamental diagram (FD) reconstruction~\cite{shi2021physics}, discrete- or continuous-time recurrent neural networks~\cite{ma2015long,liang2021modeling}, and physics-informed machine learning, which combines model- and data-based formulations~\cite{barreau2021physics,wilkman2025online}. The physics-informed strategies are particularly flexible and can fuse measurements from many different sources~\cite{di2023physics}.

These algorithms rely on sensor data. Inflow and outflow measurements from loop detectors have been used for decades to reconstruct highway traffic density~\cite{ferrara2018freeway}. More recently, probe and connected or automated vehicles have been investigated to provide vehicle-level measurements from within the traffic stream for more accurate reconstructions~\cite{lee2025traffic,baloul2025traffic}.

The rapid adoption of probe vehicles in the research community is partly motivated by their lower installation costs compared with standard sensors. However, collecting and processing their data in real time requires a complex communication architecture~\cite{herrera2010evaluation}. Here, we propose using a single probe vehicle, whose position is known over time, together with sensors already installed for speed enforcement, to reconstruct traffic conditions and dynamics along road segments. These segments are delimited by automatic number-plate recognition (ANPR) cameras, which record entry/exit timestamps (EET) for each vehicle~\cite{oliveira2012online,securiteroutiere_radarsfixes}. These timestamps contain travel-time information, which we leverage to infer the traffic dynamics.

In this paper, we use Neural Ordinary Differential Equations (Neural ODEs)~\cite{chen2018neural} to learn the speed--density relation of a follow-the-leader model and to reconstruct each vehicle's trajectory from EET data and the trajectory of a single probe vehicle. Our contributions are threefold: (i) we adapt Neural ODE training to EET measurements, where each vehicle is only observed when it enters and exits the road segment; (ii) we propose a speed--density network that satisfies the physical constraints by construction, combining a normalized monotone network with a learned regime change, while retaining the ability to approximate any admissible relation; and (iii) we validate the method on first- and second-order traffic simulations and on real-world ring-road data~\cite{stern2018dissipation}. After introducing the preliminaries and the problem in Sections~\ref{sec:preliminaries} and~\ref{sec:problem_statement}, we present (i) and (ii) in Section~\ref{sec:method} and (iii) in Section~\ref{sec:results}, and conclude in Section~\ref{sec:conclussion}.

%% file: Sections/PreliminariesProblemStatement.tex
\section{Preliminaries} 
\label{sec:preliminaries}

Before stating the problem, we cover fundamentals of microscopic traffic flow modeling, Neural ODEs, and constrained neural networks to clearly introduce our contribution.

\subsection{Microscopic Traffic Flow Modeling}
The most granular approach to modeling traffic flow is microscopic: each vehicle has its own trajectory, computed as the solution to an ordinary differential equation~\cite[Section~5.2]{ferrara2018freeway}. Under the first-order assumption, the speed of vehicle $i \in \{1, \dots, N\}$ depends directly on the local density $\rho_i(t)$ around it, which leads to the follow-the-leader model $\dot x_i(t) = v(\rho_i(t))$, $x_i(0) = x_i^0$, where $x_i$ is the position of vehicle $i$ and $x_i^0$ its initial position. The global density is the number of vehicles on the road divided by the road's length, and is proportional to the occupancy ratio; the local density is its limit as the length of the road segment approaches $0$. If $x_0^0 > x_1^0 > \cdots > x_N^0$, the local density around vehicle $i$ is approximated by:
\begin{equation}
    \label{eq:rho}
    \rho_i(t) \simeq \frac{L_{car}}{x_{i-1}(t) - x_i(t)},
\end{equation}
where $L_{car}$ is the length of a vehicle plus the minimum spacing between vehicles~\cite{di2017follow}. The density is thereby normalized: $\rho_i \to 0$ as $x_{i-1} - x_i \to \infty$, and $\rho_i \to 1$ as $x_{i-1} - x_i \to L_{car}$, i.e., when vehicles are bumper-to-bumper. The follow-the-leader model then becomes:
\begin{equation}
    \label{eq:follow_the_leader}
    \begin{cases}
        \dot{x}_i(t) = v\left( \frac{L_{car}}{x_{i-1}(t) - x_i(t)} \right), \quad& x_i(0) = x_i^0, \\
        x_0 \in \mathcal{C}(\mathbb{R}_+, \mathbb{R}_+),
    \end{cases}
\end{equation}
where $x_0$ is the known trajectory of the leader vehicle. 

The speed--density relation $v$ derives from the fundamental diagram~\cite[Section~2.1.3]{ferrara2018freeway} and is usually partially unknown and stochastic~\cite{qu2017stochastic}. However, some fundamental properties can be inferred. First, when the road is almost empty, i.e., $\rho \approx 0$, vehicles move at the free-flow speed $v_{free}$, independently of their neighbors, as in the free-flow regime~\cite{kerner2009introduction}. Second, when vehicles are bumper-to-bumper, i.e., $\rho \approx 1$, they cannot move without colliding. Above a critical density, vehicles are slowed down by those ahead; this congested regime is much more complex to model. Furthermore, on average, the speed of a vehicle decreases with the local density~\cite{barreau2021physics}: the less congested the road, the higher the speed. We summarize these properties of $v$ as:
\begin{subequations}
    \label{eq:constraints}
    \begin{align}
        v(0) &= v_{free}, \\
        v(1) &= 0, \\
        v(\rho_1) &\geq v(\rho_2), \quad 0 \leq \rho_1 \leq \rho_2 \leq 1.
    \end{align}
\end{subequations}
These observations have led to various models of the steady-state homogeneous behavior. Greenshields~\cite{greenshields1935study} proposed the linear model $v(\rho) = v_{free}(1 - \rho)$, which has convenient analytical properties but does not capture the regime transition. Newell and Daganzo~\cite{NEWELL1993281,DAGANZO1994269} later proposed a model that accounts for the regime transition through a triangular flow--density relation $q(\rho) = \rho v(\rho)$, corresponding to the speed--density relation
\begin{equation}
    \label{eq:newell}
    v_{ND}(\rho) = v_{free} \min\left(1,\, w\left(\tfrac{1}{\rho} - 1\right)\right),
\end{equation}
with $w > 0$, which has a kink at the critical density $\rho_c = w/(1+w)$ separating the free-flow and congested regimes. Regardless of the model, the congested regime is difficult to capture because factors such as reaction times can cause different speeds to correspond to the same density, suggesting that a first-order model, where speed adapts instantaneously to density, is too crude an approximation of real-world traffic~\cite{delle2021three}.

Second-order models avoid instantaneous velocity changes by modeling the acceleration instead of the speed, based on the vehicle's own speed, the spacing $s_i(t) = x_{i-1}(t) - x_i(t)$, and the relative speed $\Delta v_i(t) = v_{i-1}(t) - v_i(t)$ to the preceding vehicle:
\begin{equation}
    \label{eq:second_order}
    \begin{cases}
        \dot{x}_i(t) = v_i(t), \quad& x_i(0) = x_i^0, \\
        \dot{v}_i(t) = a\left(v_i(t), s_i(t), \Delta v_i(t) \right), \quad& v_i(0) = v_i^0, \\
        x_0, v_0 \in \mathcal{C}(\mathbb{R}_+, \mathbb{R}_+),
    \end{cases}
\end{equation}
where $x_0, v_0$ are the known trajectory and velocity of the leader vehicle, and the acceleration function $a$ is continuous. Although they do not capture the inherent stochasticity of real-world traffic, second-order models reproduce more realistic behavior, since the speed can deviate from the fundamental diagram while staying close to it on average~\cite[Section~5.2.1]{ferrara2018freeway}. The downside is that without careful modeling, collisions may occur because the speed can remain nonzero as the gap approaches $0$. Among the available second-order models, the Intelligent Driver Model (IDM)~\cite{treiber2000congested} combines a free-road term, which drives the vehicle toward its desired speed, with an interaction term that maintains a safe gap to the leader and anticipates its speed changes, thereby avoiding collisions.

\subsection{Learning Dynamics with Neural ODEs}
\label{sec:neuralode}

We define a multi-layer perceptron (MLP)~\cite{Goodfellow-et-al-2016} with $l_\phi \geq 1$ hidden layers of $n_\phi$ neurons each as
\begin{equation}
    \label{eq:MLP}
    \phi(\rho; \theta) := W_{l_\phi+1} \big(H_{l_\phi} \circ \cdots \circ H_1\big)(\rho) + b_{l_\phi+1},
\end{equation}
where $H_i(z) := \sigma(W_i z + b_i)$ is the $i$-th hidden layer with an element-wise non-linear activation function $\sigma$, and $\theta = \{(W_i, b_i)\}_{i=1}^{l_\phi+1}$ collects the weights $W_1 \in \mathbb{R}^{n_\phi \times 1}$, $W_{l_\phi+1} \in \mathbb{R}^{1 \times n_\phi}$, and $W_i \in \mathbb{R}^{n_\phi \times n_\phi}$ otherwise, and the biases $b_{l_\phi+1} \in \mathbb{R}$ and $b_i \in \mathbb{R}^{n_\phi}$ otherwise. With enough neurons, an MLP can approximate any continuous function on a compact set arbitrarily well~\cite{hornik1991approximation}.

Consider a dynamical system with state $x(t) \in \mathbb{R}^D$,
\begin{equation}
    \label{eq:generic_dynamics}
    \dot x(t) = f\big(t, x(t)\big), \qquad x(0) = x^0,
\end{equation}
whose dynamics $f$ are partially unknown, such as the speed--density relation in~\eqref{eq:follow_the_leader} or the acceleration in~\eqref{eq:second_order}. A Neural ODE~\cite{chen2018neural} replaces $f$ with trainable dynamics $\hat f(\cdot, \cdot; \theta)$, typically built from an MLP with parameters $\theta$, and predicts the state by integrating them from the same initial condition:
\begin{equation}
    \label{eq:reconstruct}
    \hat x(t; \theta) = x^0 + \int_0^t \hat f\big(\tau, \hat x(\tau;\theta); \theta\big)\,d\tau,
\end{equation}
where the integral is computed with a numerical solver such as the fourth-order Runge--Kutta method (RK4). The parameters $\theta$ are found by minimizing a loss that compares $\hat x$ with measurements of $x$, which may be available only at a few time instants, e.g., the terminal cost $\Phi(\hat x(T;\theta)) = \|\hat x(T;\theta) - x(T)\|^2$. Learning the dynamics rather than the trajectory generalizes beyond the sampled points, since the model captures the mechanism that generates the data. It also allows known physics to be built into $\hat f$ by adding a learned correction to a known physical model~\cite{rackauckas2020universal} or, as in this paper, by constraining its structure.

Computing the gradient of the loss by backpropagating through the solver requires storing every intermediate state, so memory grows with the number of solver steps, limiting the time window of the problem. The adjoint sensitivity method~\cite{pontryagin2018mathematical, chen2018neural} avoids this by solving a second ODE backward in time, and extends to losses that also contain a running cost $\ell$~\cite{miao2025opt}:
\begin{equation}
    \label{eq:miao_loss}
    \mathcal{L}(\theta) = \Phi\big(\hat x(T; \theta)\big) + \int_0^T \ell\big(t, \hat x(t; \theta)\big)\,dt.
\end{equation}
The gradient is
\begin{equation}
    \label{eq:adjoint_gradient}
    \nabla_\theta \mathcal{L}(\theta) = \int_0^T \left(\frac{\partial \hat f}{\partial \theta}\right)^{\!\top} \lambda(t)\,dt,
\end{equation}
where the adjoint $\lambda$ is obtained by solving, backward in time,
\begin{equation}
    \label{eq:miao_dynamics}
    \begin{aligned}
        \dot{\hat x}(t) &= \hat f\big(t, \hat x(t);\theta\big), \qquad \hat x(T) \text{ given by~\eqref{eq:reconstruct}}, \\
        \dot\lambda(t) &= -\left(\frac{\partial \hat f}{\partial \hat x}\right)^{\!\top} \lambda(t) - \left(\frac{\partial \ell}{\partial \hat x}\right)^{\!\top}, \\
        \lambda(T) &= \left(\frac{\partial \Phi}{\partial \hat x}\big(\hat x(T)\big)\right)^{\!\top}.
    \end{aligned}
\end{equation}
All partial derivatives are evaluated at $(t, \hat x(t;\theta);\theta)$, and the dependence of $\hat x$ on $\theta$ is omitted for brevity. The gradient~\eqref{eq:adjoint_gradient} is then passed to a gradient-based optimizer, such as Adam~\cite{kingma2014adam}, to update $\theta$.

\subsection{Constrained Neural Networks}
\label{sec:constrained_nn}
One way to incorporate prior knowledge into an MLP is to impose constraints through a tailored architecture, ensuring they hold for all parameter values. Boundary and endpoint conditions are commonly enforced this way~\cite{lagaris1998artificial, chen2023hard}. For the endpoint conditions in~\eqref{eq:constraints}, \cite{wilkman2025online} multiplies the MLP~\eqref{eq:MLP} by a factor that vanishes at $\rho = 0$ and $\rho = 1$, and adds a term that equals $v_{free}$ at $\rho = 0$ and $0$ at $\rho = 1$:
\begin{equation}
    \label{eq:constraint_old}
    \tilde\phi(\rho;\theta) := \rho(1-\rho)\,\phi(\rho;\theta) + (1-\rho)\,v_{free},
\end{equation}
which we refer to as the fixed-endpoint model. Then $\tilde\phi(0;\theta) = v_{free}$ and $\tilde\phi(1;\theta) = 0$ for any $\theta$. The main benefit of such hard constraints is reliability: the learned relation satisfies the physical properties regardless of whether the data contains information about the endpoints or whether the training is perturbed by noise. However,~\eqref{eq:constraint_old} does not enforce monotonicity, so $\tilde\phi$ may exceed $v_{free}$ or become negative inside $(0,1)$.

Monotonicity can also be enforced by construction: if all weights $W_i$ in~\eqref{eq:MLP} are positive and $\sigma$ is strictly increasing, such as $\tanh$, then $\phi$ is strictly increasing, and a decreasing network is obtained by negating its output. Such networks can approximate any continuous non-decreasing function on $[0,1]$ arbitrarily well~\cite{daniels2010monotone}. However, $\tilde\phi$ cannot simply be combined with a monotonicity construction. First, a monotone $\phi$ does not make $\tilde\phi$ monotone: a constant $\phi \equiv c > v_{free}$, for instance, makes $\tilde\phi$ increase near $\rho = 0$. Second, a monotone $\phi$ cannot represent every admissible relation: the $\phi$ for which $\tilde\phi$ equals the Newell--Daganzo relation increases below the critical density and decreases just above it. Section~\ref{sec:method} proposes a construction that enforces both properties while retaining universal approximation.

\section{Problem statement}
\label{sec:problem_statement}
We consider a single-lane road segment $[x_e, x_f]$ which is traversed by a leading probe vehicle with trajectory $t \mapsto x_0(t)$ and $N$ trailing vehicles $x_i$, $i \in \{1, \dots, N\}$, with the ordered positions $x_i(t) > x_{i+1}(t)$ for all $t \geq 0$ (vehicles do not overtake each other). 

\begin{figure}[t]
    \centering
    \includegraphics[width=0.95\linewidth]{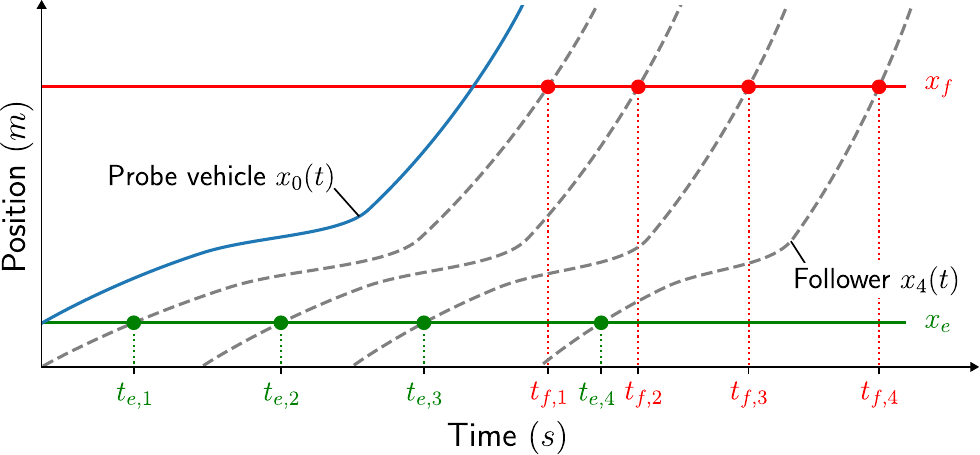}
    \caption{Problem setting with $N = 4$ followers. The probe vehicle trajectory $x_0(t)$ (solid) is known, while the follower trajectories $x_i(t)$ (dashed) are unknown; only their entry times $t_{e,i}$ at $x_e$ and exit times $t_{f,i}$ at $x_f$ (dots) are measured.}
    \label{fig:data_demo}
\end{figure}

We assume that the road segment is delimited by entry and exit cameras which record the time when each vehicle enters and exits the segment \cite{oliveira2012online}. This leads to the dataset
\begin{equation}
    \mathcal{D} = \left\{(t_{e,i}, t_{f,i})\right\}_{i=1}^N
\end{equation}
where $x_i(t_{e,i}) = x_e$ and $x_i(t_{f,i}) = x_f$. We also assume knowledge of the probe vehicle, which continuously communicates its position $x_0(t)$ to the infrastructure. Fig.~\ref{fig:data_demo} illustrates the setting. Leveraging this data, we have two objectives: (i) to identify the speed--density relation and (ii) to reconstruct the vehicles' trajectories.

To achieve these goals, we model the vehicles with the follow-the-leader model~\eqref{eq:follow_the_leader}, where the speed--density relation is a neural network $\rho \mapsto \hat v(\rho; \theta)$ with parameters $\theta$, and the reconstructed trajectories $\hat x_i(t;\theta)$ are obtained by integrating $\hat v$. The two objectives amount to finding parameters $\theta$ that minimize the trajectory error
\begin{equation}
    \label{eq:traj_loss}
    \mathcal{L}_{x}(\theta) := \frac{1}{N}\sum_{i=1}^N \int_{t_{e,i}}^{t_{f,i}} |\hat x_i(t;\theta) - x_i(t)|^2\,dt,
\end{equation}
and the speed--density error along the true trajectories,
\begin{equation} 
    \label{eq:vel_loss}
    \mathcal{L}_{v}(\theta) := \frac{1}{N}\sum_{i=1}^N \int_{t_{e,i}}^{t_{f,i}} |\hat v(\rho_i(t);\theta) - \dot x_i(t)|^2\, dt,
\end{equation}
where $\rho_i(t)$ is the local density~\eqref{eq:rho} of the true trajectories. Since~\eqref{eq:traj_loss} averages over time and vehicles, we also consider the worst-case trajectory error, i.e., the maximum deviation
\begin{equation}
    \label{eq:max_dev}
    \mathcal{L}_{\max}(\theta) := \max_{i \in \{1, \dots, N\}}\ \max_{t \in [t_{e,i}, t_{f,i}]} |\hat x_i(t;\theta) - x_i(t)|.
\end{equation}
However, all three errors depend on the trajectories inside the road segment, which are not observed; instead, we minimize the exit error, which only requires $\mathcal{D}$ and $x_0$:
\begin{equation}
    \label{eq:loss}
    \mathcal{L}(\theta) = \frac{1}{N}\sum_{i=1}^N (\hat x_i(t_{f,i};\theta) - x_f)^2.
\end{equation}
This problem is a Neural ODE as described in Section~\ref{sec:neuralode}. As we have $\mathcal{L}_x(\theta) = 0 \Rightarrow \mathcal{L}(\theta) = 0$, but not the converse, we want to incorporate as much physical knowledge as possible by enforcing the traffic flow theory constraints \eqref{eq:constraints}, which narrows the set of possible solutions to be consistent with the data. The next section proposes an adapted training algorithm to effectively minimize \eqref{eq:loss} and enforce the constraints natively in $\hat{v}$.

%% file: Sections/Method.tex
\section{Method}
\label{sec:method}
This section first adapts Neural ODE training to EET measurements, and then constructs a speed--density network that satisfies the constraints~\eqref{eq:constraints} by design and captures the regime change at the critical density.

\subsection{Neural ODE for EET Measurements}
\label{sec:node_training}
Compared with the setting of Section~\ref{sec:neuralode}, the followers are coupled through their densities, and vehicle $i$ is only observed at its entry and exit times $t_{e,i}$ and $t_{f,i}$, with no information about its position before $t_{e,i}$. We therefore keep each follower at the entry position until it enters the section, then let it follow the learned dynamics afterward. Denoting the estimated positions of the followers by $\hat x(t;\theta) = (\hat x_1(t;\theta), \dots, \hat x_N(t;\theta))^\top$ and setting $\hat x_0(t;\theta) := x_0(t)$, the known probe vehicle trajectory, the estimated dynamics of follower $i \in \{1, \dots, N\}$ are
\begin{equation}
    \label{eq:entry_dynamics}
    \frac{d\hat x_i(t;\theta)}{dt} = \hat f_i(t, \hat x;\theta) := \begin{cases} \hat v\big(\hat \rho_i(t);\theta\big), & t \geq t_{e,i}, \\ 0, & \text{otherwise}, \end{cases}
\end{equation}
with $\hat x_i(0;\theta) = x_e$, where $\hat\rho_i(t) := L_{car}/(\hat x_{i-1}(t;\theta) - \hat x_i(t;\theta))$ is the local density~\eqref{eq:rho} evaluated along the estimated trajectories. Since vehicles do not overtake, vehicle $i-1$ enters before vehicle $i$, so $\hat\rho_i$ only needs to be defined for $t \geq t_{e,i}$. The stacked right-hand side $\hat f := (\hat f_1, \dots, \hat f_N)^\top$ is the trainable dynamics of the Neural ODE~\eqref{eq:reconstruct}. Since a vehicle that has exited still affects the vehicles behind it, all trajectories are integrated up to the last exit time $T := t_{f,N}$.

The exit error~\eqref{eq:loss} evaluates each follower at its own exit time rather than at a single terminal time. It therefore takes the form~\eqref{eq:miao_loss} with
\begin{align}
    \label{eq:terminal}
    \Phi(\hat x(T;\theta)) &:= \frac{1}{N}\big(\hat x_N(T;\theta) - x_f\big)^2, \\
    \label{eq:running}
    \ell(t, \hat x(t;\theta)) &:= \frac{1}{N}\sum_{i=1}^{N-1} \big(\hat x_i(t;\theta) - x_f\big)^2\, \delta(t - t_{f,i}),
\end{align}
where $\delta$ is the Dirac delta, i.e., $\int_0^T \varphi(t)\,\delta(t - \tau)\,dt = \varphi(\tau)$ for $\tau \in (0, T)$ and any $\varphi$ continuous at $\tau$. Inserting~\eqref{eq:terminal} and~\eqref{eq:running} into~\eqref{eq:miao_dynamics}, the adjoint starts from $\lambda_N(T) = \frac{2}{N}(\hat x_N(T;\theta) - x_f)$ and $\lambda_i(T) = 0$ for $i < N$, follows the homogeneous adjoint dynamics between exit times, and jumps at each exit time:
\begin{equation}
    \label{eq:jump}
    \lambda_i(t_{f,i}^-) = \lambda_i(t_{f,i}^+) + \frac{2}{N}\big(\hat x_i(t_{f,i};\theta) - x_f\big)
\end{equation}
for $i \in \{1, \dots, N-1\}$, while the other components are continuous. The gradient then follows from Section~\ref{sec:neuralode}; since the dynamics of follower $i$ vanish before $t_{e,i}$, it only contributes to the gradient after it has entered. Since $\hat \rho_{i+1}$ depends on $\hat x_i$, vehicle $i$ still influences its follower after exiting. This is why the trajectories are integrated up to time $T$, and why the probe trajectory must be known until the last exit.

\subsection{Enforcing Traffic Dynamics with Hard Constraints}
\label{sec:hard_constraints}
As noted in Section~\ref{sec:problem_statement}, a vanishing exit error does not imply a vanishing trajectory error: there may be several speed--density relations that reproduce the entry and exit times equally well, and the available data alone cannot distinguish between the solutions. To help learning and prevent unrealistic learned models, we restrict $\hat v$ to physically admissible relations, i.e., relations satisfying~\eqref{eq:constraints}~\cite{barreau2021physics}. A continuous $v$ satisfies~\eqref{eq:constraints} if and only if $v/v_{free}$ belongs to the set of normalized admissible speed--density relations:
\begin{equation}
    \label{eq:admissible}
    \mathcal{A} := \left\{ g \in \mathcal{C}([0,1]) : \begin{aligned} & g \text{ non-increasing}, \\ & g(0) = 1,\ g(1) = 0 \end{aligned} \right\},
\end{equation}
where $\mathcal{C}([0,1])$ denotes the continuous real functions on $[0,1]$, equipped with the norm $\|\cdot\|_\infty$. We construct $\hat v$ such that $\hat v/v_{free} \in \mathcal{A}$ for every $\theta$ from three components: a monotone network (Section~\ref{sec:monotone}), a normalization that fixes the endpoints (Section~\ref{sec:endpoints}), and a regime change at a critical density (Section~\ref{sec:regime}), which are combined in Section~\ref{sec:final_model}.

\subsubsection{Enforcing Monotonicity}
\label{sec:monotone}
As noted in Section~\ref{sec:constrained_nn}, an MLP with positive weights and strictly increasing activation functions is strictly increasing. Parameterizing the weights by their absolute value enforces positivity but is not differentiable at $0$, which may cause issues during training; we instead use the smooth reparameterization
\begin{equation}
    \label{eq:positive}
    (\mathcal{P}_\epsilon\phi)(\rho;\theta) := \phi(\rho;\theta_\epsilon), \quad \theta_\epsilon := \left\{\left(\sqrt{W_i^{2} + \epsilon},\, b_i\right)\right\}_{i=1}^{l_\phi+1},
\end{equation}
where the square and the square root are applied elementwise and $\epsilon > 0$. The weights of $\mathcal{P}_\epsilon\phi$ are bounded below by $\sqrt{\epsilon}$, so with $\sigma = \tanh$, $\mathcal{P}_\epsilon\phi$ is strictly increasing for every $\theta$. Conversely, any MLP whose weights $\tilde W_i$ are all at least $\sqrt{\epsilon}$ equals $\mathcal{P}_\epsilon\phi$ with $W_i = \sqrt{\tilde W_i^2 - \epsilon}$ elementwise. Hence, as $\epsilon \to 0$, $\mathcal{P}_\epsilon\phi$ can represent any MLP with positive weights, and thereby approximate any continuous non-decreasing function on $[0,1]$ arbitrarily well (Section~\ref{sec:constrained_nn}). However, $\mathcal{P}_\epsilon\phi$ is increasing rather than decreasing and does not satisfy the endpoint conditions; the normalization introduced next corrects both issues.

\subsubsection{Fixed Endpoints}
\label{sec:endpoints}
Since~\eqref{eq:constraint_old} cannot be combined with a monotone network (Section~\ref{sec:constrained_nn}), we instead normalize the network output by its own endpoint values. For $\phi \in \mathcal{C}([0,1])$ with $\phi(0) \neq \phi(1)$, we define the operator:
\begin{equation}
    \label{eq:normalization}
    (\mathcal{T}\phi)(\rho) := \frac{\phi(\rho) - \phi(1)}{\phi(0) - \phi(1)}, \qquad \rho \in [0,1].
\end{equation}
By construction, $\mathcal{T}\phi$ is continuous with $(\mathcal{T}\phi)(0) = 1$ and $(\mathcal{T}\phi)(1) = 0$, and since $g(0) - g(1) = 1$ and $g(1) = 0$ for $g \in \mathcal{A}$, $\mathcal{T}$ leaves admissible relations unchanged, i.e., $\mathcal{T}g = g$. Moreover, $\mathcal{T}$ is invariant to affine transformations of its argument: for any scalars $a \neq 0$ and $b$, $\mathcal{T}(a\phi + b) = \mathcal{T}\phi$, since $b$ cancels and $a$ factors out of both the numerator and the denominator of~\eqref{eq:normalization}.

\begin{theorem}
    \label{thm:T}
    Let $\phi \in \mathcal{C}([0,1])$ with $\phi(0) \neq \phi(1)$. Then:
    \begin{enumerate}
        \item[(i)] if $\phi$ is non-decreasing or non-increasing, then $\mathcal{T}\phi \in \mathcal{A}$;
        \item[(ii)] if $\eta := \|\phi - g\|_\infty < \frac{1}{2}$ for some $g \in \mathcal{A}$, then $\|\mathcal{T}\phi - g\|_\infty \leq \frac{2\eta}{1 - 2\eta}\xrightarrow[\eta \to 0]{} 0$.
    \end{enumerate}
\end{theorem}

\begin{proof}
    (i) For $\rho_1 \leq \rho_2$, $(\mathcal{T}\phi)(\rho_1) - (\mathcal{T}\phi)(\rho_2) = \big(\phi(\rho_1) - \phi(\rho_2)\big)/\big(\phi(0) - \phi(1)\big)$. If $\phi$ is non-increasing, the numerator is non-negative and the denominator positive; if $\phi$ is non-decreasing, the numerator is non-positive and the denominator negative. In both cases, $\mathcal{T}\phi$ is non-increasing, and since it is continuous with $(\mathcal{T}\phi)(0) = 1$ and $(\mathcal{T}\phi)(1) = 0$, $\mathcal{T}\phi \in \mathcal{A}$.

    (ii) Let $e := \phi - g$, so that $|e(\rho)| \leq \eta$ for all $\rho$. Since $g(0) = 1$ and $g(1) = 0$, the denominator of~\eqref{eq:normalization} is $d := \phi(0) - \phi(1) = 1 + e(0) - e(1)$, so $|d| \geq 1 - 2\eta > 0$. Using $\phi = g + e$,
    \begin{equation*}
        \big((\mathcal{T}\phi)(\rho) - g(\rho)\big)\, d = e(\rho) - \big(1 - g(\rho)\big) e(1) - g(\rho)\, e(0).
    \end{equation*}
    Since $0 \leq g(\rho) \leq 1$, the right-hand side is bounded in absolute value by $\eta\big(1 + (1 - g(\rho)) + g(\rho)\big) = 2\eta$, and dividing by $|d| \geq 1 - 2\eta$ gives the bound.
\end{proof}

By Theorem~\ref{thm:T}(i), $\mathcal{T}$ maps any strictly monotone function to an admissible relation, and by Theorem~\ref{thm:T}(ii), it preserves approximation accuracy up to a factor of about two; Section~\ref{sec:final_model} combines both properties in the proposed model.

\subsubsection{Regime Change}
\label{sec:regime}
Empirical speed--density data show a free-flow and a congested regime separated by a critical density, at which the slope of the relation changes abruptly, as in the Newell--Daganzo relation~\eqref{eq:newell}. With smooth activation functions such as $\tanh$, $\mathcal{T}\mathcal{P}_\epsilon\phi$ needs many neurons to approximate such a kink. Instead, we build the kink into the model by rectifying the increasing network $\psi := \mathcal{P}_\epsilon\phi$ before normalizing it. If $\psi(0) \leq 0 < \psi(1)$,
\begin{equation}
    \label{eq:regime}
    \big(\mathcal{T}\max(0, \psi)\big)(\rho) = 1 - \frac{\max\big(0, \psi(\rho)\big)}{\psi(1)},
\end{equation}
which equals $1$ up to the critical density $\hat\rho_c$ at which $\psi(\hat\rho_c) = 0$ and strictly decreases after it, giving a free-flow plateau followed by a congested branch. The critical density is thus set by the network itself and learned with it, without additional parameters.

However, with $\max(0, \cdot)$,~\eqref{eq:regime} is exactly $1$ below $\hat\rho_c$, so its gradient with respect to $\theta$ vanishes there, such that vehicles in the free-flow regime do not contribute to training, and training stalls if all estimated densities are below $\hat\rho_c$. Moreover, $\mathcal{T}\max(0, \psi)$ is undefined if $\psi(1) \leq 0$, as then $\max(0, \psi(0)) = \max(0, \psi(1)) = 0$. We therefore replace $\max(0, \cdot)$ by the softplus function
\begin{equation}
    \label{eq:critical}
    \mathcal{S}_\beta(z) := \frac{1}{\beta}\log\left(1 + e^{\beta z}\right),
\end{equation}
a smooth, positive and strictly increasing approximation of $\max(0, z)$ with sharpness $\beta > 0$, satisfying $0 < \mathcal{S}_\beta(z) - \max(0, z) \leq \log(2)/\beta$ for all $z$, so the error vanishes as $\beta \to \infty$. Applied pointwise, $\mathcal{S}_\beta\psi$ is positive and strictly increasing for every $\theta$, so $\mathcal{T}\mathcal{S}_\beta\psi$ is always well defined, and its gradient no longer vanishes below $\hat\rho_c$. Moreover, $\mathcal{S}_\beta(cz) = c\,\mathcal{S}_{c\beta}(z)$ for $c > 0$, so by the affine invariance of $\mathcal{T}$,
\begin{equation}
    \label{eq:sharpness}
    \mathcal{T}\mathcal{S}_\beta(c\psi) = \mathcal{T}\big(c\,\mathcal{S}_{c\beta}\psi\big) = \mathcal{T}\mathcal{S}_{c\beta}\psi.
\end{equation}
Scaling the network output by $c$ is thus equivalent to sharpening the softplus by the same factor, so the network can sharpen the kink itself, the sharpness is effectively learned, and $\beta$ mainly sets its value at initialization. In practice, $\mathcal{S}_\beta\mathcal{P}_\epsilon\phi$ is simply a monotone network with a softplus output activation. Finally, the rectification does not reduce expressiveness: for any $g \in \mathcal{A}$, the increasing function $1 - g$ is non-negative and hence unchanged by $\max(0, \cdot)$, and since $\mathcal{T}$ removes the output scaling in~\eqref{eq:sharpness}, the softplus error can be made arbitrarily small for any fixed $\beta$ (Proposition~\ref{prop:approx}).

\subsubsection{Proposed Model}
\label{sec:final_model}
Combining the monotone network, the regime change, and the normalization, the proposed model is
\begin{equation}
    \label{eq:final_model}
    \hat v(\rho;\theta) := v_{free}\,\big(\mathcal{T}\mathcal{S}_\beta\mathcal{P}_\epsilon\phi\big)(\rho;\theta).
\end{equation}
Since $\mathcal{S}_\beta\mathcal{P}_\epsilon\phi$ is strictly increasing, Theorem~\ref{thm:T}(i) gives $\hat v/v_{free} \in \mathcal{A}$, i.e., $\hat v$ satisfies~\eqref{eq:constraints} for every $\theta$, $\epsilon > 0$, and $\beta > 0$, and since $\mathcal{T}$ is applied last, the endpoint conditions hold exactly. Further, Proposition~\ref{prop:approx} shows that the proposed model can approximate any admissible relation arbitrarily well for every fixed $\beta$.

\begin{proposition}
    \label{prop:approx}
    Let $g \in \mathcal{A}$ and $\beta > 0$. For every $\delta > 0$ and every sufficiently small $\epsilon > 0$, there exists an MLP $\phi$ such that $\|\mathcal{T}\mathcal{S}_\beta\mathcal{P}_\epsilon\phi - g\|_\infty \leq \delta$.
\end{proposition}

\begin{proof}
    Let $h := 1 - g$, which is continuous, non-decreasing, and non-negative, and let $\gamma > 0$ and $c > 0$. Since $ch$ is continuous and non-decreasing, for a sufficiently wide $\phi$ and sufficiently small $\epsilon$, there exists $\theta$ such that $\psi := \mathcal{P}_\epsilon\phi$ satisfies $\|\psi - ch\|_\infty \leq c\gamma$ (Section~\ref{sec:monotone}), and by~\eqref{eq:sharpness}, $\mathcal{T}\mathcal{S}_\beta\psi = \mathcal{T}\mathcal{S}_{c\beta}(\psi/c)$. Since $\mathcal{S}_{c\beta}$ is $1$-Lipschitz and $h = \max(0, h)$, so that $0 < \mathcal{S}_{c\beta}(h) - h \leq \log(2)/(c\beta)$, the triangle inequality gives
    \begin{equation*}
        \|\mathcal{S}_{c\beta}(\psi/c) - h\|_\infty \leq \gamma + \log(2)/(c\beta) =: \eta.
    \end{equation*}
    By the affine invariance of $\mathcal{T}$, $\mathcal{T}\mathcal{S}_{c\beta}(\psi/c) = \mathcal{T}\big(1 - \mathcal{S}_{c\beta}(\psi/c)\big)$, where $\|1 - \mathcal{S}_{c\beta}(\psi/c) - g\|_\infty = \|\mathcal{S}_{c\beta}(\psi/c) - h\|_\infty \leq \eta$. Hence, if $\eta < 1/2$, Theorem~\ref{thm:T}(ii) gives
    \begin{equation*}
        \|\mathcal{T}\mathcal{S}_\beta\mathcal{P}_\epsilon\phi - g\|_\infty \leq \frac{2\eta}{1 - 2\eta},
    \end{equation*}
    which is at most $\delta$ if $\eta \leq \delta/(2 + 2\delta)$, e.g., for $\gamma \leq \delta/(4 + 4\delta)$ and $c \geq (4 + 4\delta)\log(2)/(\beta\delta)$.
\end{proof}

%% file: Sections/ResultsDiscussion.tex
\section{Numerical Results}
\label{sec:results}

\begin{figure*}[!t] 
    \centering
    \includegraphics[width=0.95\linewidth]{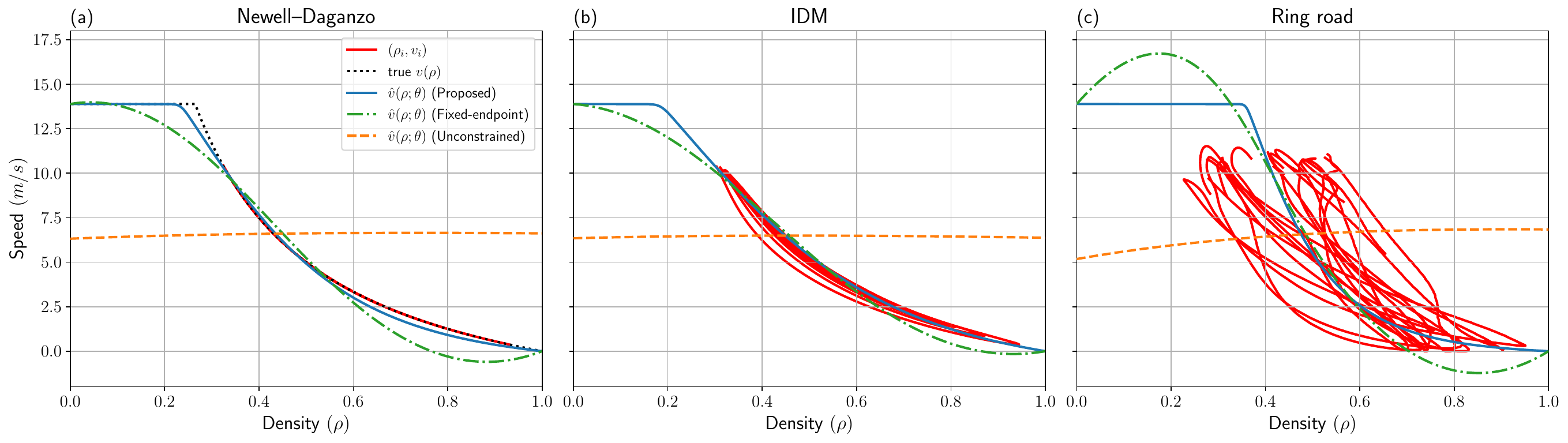}
    \caption{Learned speed--density relations $\hat v(\rho;\theta)$ from one of the five training runs in Table~\ref{tab:cases} for (a) Newell--Daganzo first-order model, (b) IDM, and (c) ring-road data for three models: proposed model (blue, solid), fixed-endpoint model~\eqref{eq:constraint_old} (green, dash-dotted), and unconstrained model (orange, dashed). The dotted black line in (a) is the true relation. Red: speed--density trajectories $(\rho_i(t), v_i(t))$ of the true followers inside the section, shown for reference only and not used in training; they cover $\rho \in [0.3, 0.95]$ in (a) and (b), and $\rho \in [0.23, 0.95]$ in (c). Within this range, the proposed model closely follows the true relation in (a) and the data in (b) and (c), except at the lowest densities in (c), where its plateau at $v_{free}$ lies above the observed speeds; the unconstrained model settles on an almost constant speed. Outside this range, the learned relation is set by the model structure alone: the unconstrained model violates all three constraints, the fixed-endpoint model becomes negative near $\rho = 1$ in all cases and exceeds $v_{free}$ by about $3$~m/s in (c), whereas the proposed model satisfies~\eqref{eq:constraints} by construction.}
    \label{fig:dynamics}
\end{figure*}
\begin{figure*}[!t]
    \centering
    \includegraphics[width=0.95\linewidth]{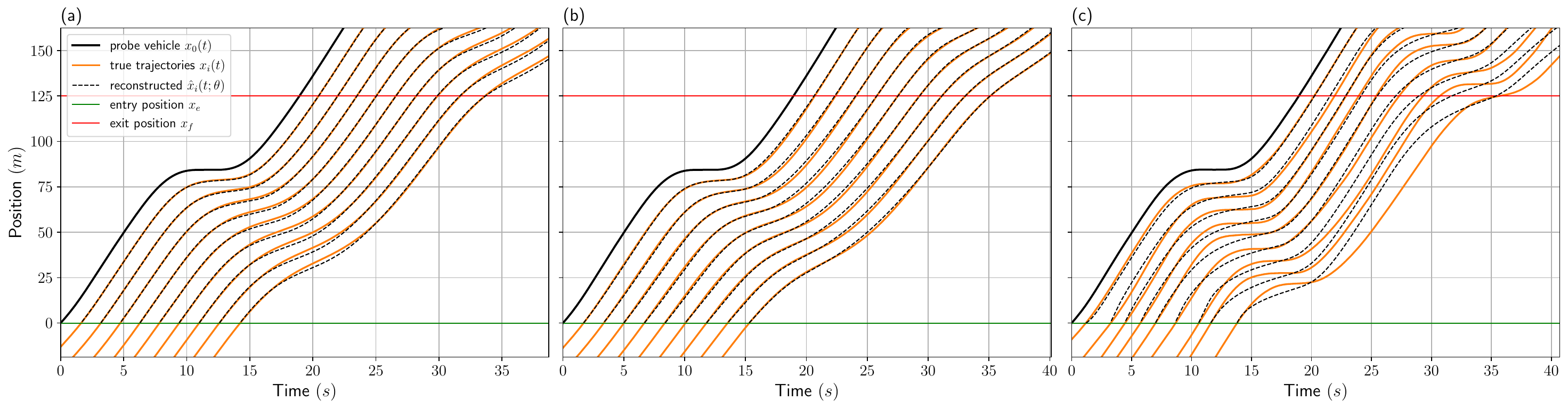}
    \caption{True (orange) and reconstructed (black, dashed) trajectories with the proposed model, from the same training run as Fig.~\ref{fig:dynamics}, for (a) Newell--Daganzo, (b) IDM, and (c) ring-road data. Each reconstruction starts at $x_e$ at the follower's entry time. The only data available are the probe vehicle's trajectory (black, solid) and the times at which the followers cross the entry and exit positions $x_e$ (green) and $x_f$ (red); the true trajectories are shown for reference only.}
    \label{fig:trajectories}
\end{figure*}

\begin{table*}[t]
  \centering
  \caption{Errors of the learned models. The exit error~\eqref{eq:loss}, which is the training loss, is evaluated at the exit times; the trajectory error~\eqref{eq:traj_loss} and the velocity error~\eqref{eq:vel_loss} are averaged over the $N = 9$ followers, and the maximum deviation~\eqref{eq:max_dev} is taken over all of them, each between the follower's entry and exit times. All but the exit error use the true trajectories, which are not available during training. Mean $\pm$ sample standard deviation over 5 training runs, each from a different random initialization and trained for 1000 iterations. The lowest mean for each scenario is in bold.}
  \label{tab:cases}
  \begin{tabular}{llcccc}
    \toprule
    Data & Model & Exit error & Trajectory error & Velocity error & Max.\ deviation \\
         &       & $\mathcal{L}$ (m$^2$) & $\mathcal{L}_x$ (m$^2\,$s) & $\mathcal{L}_v$ (m$^2$/s) & $\mathcal{L}_{\max}$ (m) \\
    \midrule
    \multirow{3}{*}{Newell--Daganzo}
      & Proposed       & {\boldmath$0.0378 \pm 0.0056$} & {\boldmath$12.6 \pm 1.6$} & {\boldmath$0.783 \pm 0.105$} & {\boldmath$2.50 \pm 0.15$} \\
      & Fixed-endpoint & $0.192 \pm 0.055$ & $54.4 \pm 12.7$ & $4.46 \pm 1.46$ & $4.70 \pm 0.44$ \\
      & Unconstrained  & $2.12 \pm 0.08$ & $1685 \pm 5$    & $155 \pm 1$     & $21.9 \pm 0.2$  \\
    \midrule
    \multirow{3}{*}{IDM}
      & Proposed       & $0.0744 \pm 0.0031$ & {\boldmath$21.9 \pm 0.3$} & {\boldmath$1.55 \pm 0.02$} & {\boldmath$2.51 \pm 0.00$} \\
      & Fixed-endpoint & {\boldmath$0.0674 \pm 0.0013$} & $45.5 \pm 2.8$  & $3.60 \pm 0.25$ & $2.99 \pm 0.10$ \\
      & Unconstrained  & $2.79 \pm 0.55$ & $1543 \pm 57$   & $128 \pm 12$    & $19.9 \pm 1.8$  \\
    \midrule
    \multirow{3}{*}{Ring road}
      & Proposed       & $4.54 \pm 0.05$ & {\boldmath$381 \pm 1$} & {\boldmath$238 \pm 2$} & $17.1 \pm 0.1$  \\
      & Fixed-endpoint & {\boldmath$4.50 \pm 0.54$} & $385 \pm 14$    & $433 \pm 130$   & {\boldmath$16.9 \pm 0.3$} \\
      & Unconstrained  & $17.3 \pm 2.0$  & $3176 \pm 53$   & $307 \pm 3$     & $27.9 \pm 0.3$  \\
    \bottomrule
  \end{tabular}
\end{table*}

We evaluate the method on three scenarios of increasing model mismatch. First, data simulated with the first-order Newell--Daganzo (ND) model, whose true speed--density relation is known and can be represented closely by the proposed model, so that the learned relation can be compared with it directly. Second, data simulated with the second-order IDM, in which vehicles respond to relative speed and cannot change speed instantaneously, so that no first-order model reproduces the trajectories exactly. Third, the real ring-road data from~\cite{stern2018dissipation}, where drivers behave differently from one another, and vehicle lengths are unknown.

In all scenarios, $N = 9$ followers traverse the section $[x_e, x_f] = [0, 125]$~m behind a probe vehicle whose velocity is the leader's measured velocity from the ring-road data. Both simulations use a free-flow speed $v_{free} = 50$~km/h ($\approx 13.9$~m/s) and a vehicle length $l = 4.6$~m, and their remaining parameters are chosen to match the ring-road data in behavior. ND~\cite{NEWELL1993281} uses a minimum gap $s_0 = 0.6$~m, and IDM~\cite{treiber2000congested} uses a minimum gap $s_0 = 1.0$~m, time headway $T_h = 1.0$~s, maximum acceleration $a_{max} = 2.5$~m/s$^2$, comfortable deceleration $b_{comf} = 2.0$~m/s$^2$, and acceleration exponent $4$. In the density~\eqref{eq:rho}, we set $L_{car} = l + s_0$, i.e., $5.2$~m for ND and $5.6$~m for IDM: since drivers keep at least the spacing $s_0$, it acts as an effective vehicle length, with $\rho = 1$ corresponding to full congestion. ND follows the triangular relation $v(\rho) = v_{free}\min\left(1, w(1/\rho - 1)\right)$ with $w = 0.36$, corresponding to a critical density $\rho_c = w/(1+w) \approx 0.26$.

For the ring road, the leader is the autonomous vehicle in Experiment~A of~\cite{stern2018dissipation}, followed by the $9$ vehicles directly behind it, from time $t = 104$~s of the experiment until the last follower exits the section at $t \approx 140$~s. We use the smoothed displacements and velocities provided with the data. We shift time and space such that the leader starts at $x_0(0) = 0$. The entry and exit times are the instants when each follower crosses $x_e$ and $x_f$, respectively. Since vehicle lengths and gaps are unknown, we use the same $L_{car} = 5.6$~m for the ring road as in the IDM simulation.

We compare three models that share the same MLP~\eqref{eq:MLP} backbone $\phi$, with $l_\phi = 3$ hidden layers of $n_\phi = 16$ neurons and using $\tanh$ as the activation function. The \textit{proposed model} in Section~\ref{sec:method} reparameterizes the weights via~\eqref{eq:positive} with $\epsilon = 10^{-8}$ to enforce monotonicity, and uses the softplus regime change~\eqref{eq:critical} with sharpness $\beta = 30$. Since initializations without a kink in $(0,1)$ performed worse in preliminary experiments, we initialize the weights and biases with the default initialization of PyTorch and redraw them until $\psi(0) < 0 < \psi(1)$, so that the initial critical density $\hat\rho_c$ lies in $(0,1)$. Over the five training runs, the initial $\hat\rho_c$ ranges from $0.08$ to $0.82$. The \textit{fixed-endpoint model} is defined in~\eqref{eq:constraint_old}, and the \textit{unconstrained model} is simply $\hat v = v_{free}\,\phi$, so that all three models use the free-flow speed $v_{free} = 13.9$~m/s. 

We integrate with RK4 with step $\Delta t = 0.1$~s, evaluating each exit position $\hat x_i(t_{f,i};\theta)$ at the exact exit time with a final partial step from the last time step before $t_{f,i}$. Since the speed--density relation is only defined on $[0,1]$, we clamp the estimated densities $\hat\rho_i$ to $[0,1]$ during integration, as the estimated spacings may fall below $L_{car}$ during training. We train for $1000$ iterations with Adam~\cite{kingma2014adam}, using a learning rate of $10^{-3}$ and decay rates $0.9$ and $0.99$ for the first- and second-moment estimates. The learning rate is constant for the first $900$ iterations and then decays to zero following a cosine schedule over the last $100$. Each model is trained five times from different random initializations.

Besides the exit error~\eqref{eq:loss}, which is the training loss, we report the trajectory error~\eqref{eq:traj_loss}, the velocity error~\eqref{eq:vel_loss}, and the maximum deviation~\eqref{eq:max_dev}. Since these metrics require the true trajectories, which are not used in training, they act as validation metrics. Once learned, the speed--density relation can also be reused to reconstruct trajectories from new entry and exit times on the same segment, as long as these cover a similar density range. 

\subsection{Learned Speed--Density Relations}

Fig.~\ref{fig:dynamics} shows the learned speed--density relations. For ND, the proposed model closely follows the true relation, including the plateau and the kink near the critical density $\rho_c \approx 0.26$, although the data only cover densities above $\rho \approx 0.3$; the plateau and the kink are thus provided by the model structure rather than by the data. For IDM, the speed--density trajectories form narrow loops because vehicles cannot change speed instantaneously, and the proposed model passes through their centers.

The fixed-endpoint model fits the observed density range reasonably well, although it smooths out the kink in (a), and since~\eqref{eq:constraint_old} does not enforce monotonicity, it becomes negative at high densities in all three cases and exceeds $v_{free}$ at low densities for the ring road. At high densities, the true speed is close to zero, and few data points are available, so a small deviation suffices to produce negative speeds. The unconstrained model converges to an almost constant speed of about $6.5$~m/s across all cases, close to the section's average speed. Such a relation already matches the exit positions within a few meters (Table~\ref{tab:cases}), and without prior knowledge, the unconstrained model does not escape this poor local minimum.

\subsection{Trajectory Reconstruction}
Fig.~\ref{fig:trajectories} shows the trajectories reconstructed with the proposed model. For ND and IDM, the reconstructed trajectories almost coincide with the true ones, with mean maximum deviations of $2.50$~m and $2.51$~m, respectively (Table~\ref{tab:cases}). For the ring road, reconstruction accuracy is high for the first followers but degrades further down the platoon, with a mean maximum deviation of $17.1$~m. Since each vehicle follows the reconstructed trajectory of its predecessor, errors accumulate along the platoon. Moreover, unlike in the simulations, the true trajectories in Fig.~\ref{fig:trajectories}(c) are unevenly spaced even before entering the section, showing that drivers keep different gaps at similar speeds. This heterogeneity, together with the unknown vehicle lengths, violates the assumption of a single speed--density relation with a common $L_{car}$, and leads to errors in the local densities~\eqref{eq:rho} from which the dynamics are learned. It also explains the large spread of the speed--density trajectories in Fig.~\ref{fig:dynamics}(c).

Table~\ref{tab:cases} confirms these observations. For ND, the proposed model reduces the trajectory error by a factor of about $4$ compared with the fixed-endpoint model and by two orders of magnitude compared with the unconstrained model, as it captures the regime change of the true relation. For IDM, the fixed-endpoint model attains the lowest exit error, yet its trajectory error is twice that of the proposed model, illustrating that a low exit error does not imply accurate trajectories. For the ring road, the fixed-endpoint model attains a slightly lower exit error and maximum deviation than the proposed model, by less than $2\%$, and a similar trajectory error, but almost twice its velocity error, which even exceeds that of the unconstrained model: it reproduces the exit times with an incorrect speed--density relation. Otherwise, the unconstrained model has the largest errors across all metrics. Both constrained models are thus limited by the model mismatch rather than by the constraints.

Overall, enforcing the endpoint conditions already improves learning substantially compared with the unconstrained model, as it rules out the trivial constant solution. Monotonicity does not necessarily improve the fit within the observed density range, but it guarantees that the learned relation remains admissible outside it, which the fixed-endpoint model fails to do in all three panels of Fig.~\ref{fig:dynamics}. Moreover, by restricting the relations that can reproduce the exit times, monotonicity makes a low exit error more indicative of accurate dynamics: for IDM and the ring road, the fixed-endpoint model attains lower exit errors than the proposed model, but around twice its velocity error. The accuracy gain for ND, in turn, likely stems from the regime change.

\subsection{Discussion}

Learning the dynamics requires the data to excite them. In our scenarios, the variations in the leader's velocity arise from naturally occurring stop-and-go waves in the ring-road experiment, which cause the followers to encounter a wide range of densities, from about $0.23$ to $0.95$. Without such variations, all vehicles would travel at a nearly constant speed and density, and the entry and exit times would only inform about a single point of the speed--density relation. In the spirit of a persistent excitation condition, the system is then not identifiable. Conversely, the relation outside the visited densities does not affect the trajectories and therefore cannot be learned from the data. There, it is determined by the model structure alone, which is why hard constraints are essential to obtain an admissible relation over the whole density range.

The ring-road results also highlight the limitations of a first-order model with homogeneous drivers: the same density can correspond to widely different speeds, and reconstruction errors grow along the platoon. 

\section{Conclusion}
\label{sec:conclussion}

We considered the problem of learning microscopic traffic dynamics and reconstructing vehicle trajectories from the entry and exit times on a road segment, as well as the trajectory of a probe vehicle leading the platoon. We adapted Neural ODE training to these EET measurements and introduced a normalization that, combined with a monotone network and a learned regime change, guarantees an admissible speed--density relation for every parameter value while retaining the ability to approximate any admissible relation. For simulated first- and second-order traffic, the proposed model recovered the speed--density relation of the first-order model and successfully reconstructed the trajectories of both, whereas the unconstrained model failed to reconstruct the FD and the trajectories. On real-world ring-road data, the proposed model still yields an admissible relation, but heterogeneous driver behaviors limit the reconstruction accuracy.

Future work includes learning individual driver behaviors and multiple vehicle classes, such as cars and trucks, while preserving the coupling between vehicles, extending the approach to second-order and stochastic models to capture the spread observed in real-world data, and deriving conditions on the excitation of the leader under which the speed--density relation is identifiable from entry and exit times.

\section*{ACKNOWLEDGMENT}
The authors used Claude (Anthropic) to revise the text in all sections based on the authors' draft, to improve language and presentation, and to check and polish the statements and proofs of Theorem~\ref{thm:T} (Section~\ref{sec:endpoints}) and Proposition~\ref{prop:approx} (Section~\ref{sec:final_model}).

%% file: Sections/references.bib
@article{chen2018neural,
  title={Neural ordinary differential equations},
  author={Chen, Ricky TQ and Rubanova, Yulia and Bettencourt, Jesse and Duvenaud, David K},
  journal={Adv. Neural Inf. Process. Syst.},
  volume={31},
  year={2018}
}

@inproceedings{greenshields1935study,
  title={A study of traffic capacity},
  author={Greenshields, B. D. and Bibbins, J. R. and Channing, W. S. and Miller, H. H.},
  booktitle={Proc. Highway Res. Board},
  volume={14},
  year={1935}
}

@article{DAGANZO1994269,
  title={The cell transmission model: A dynamic representation of highway traffic consistent with the hydrodynamic theory},
  author={Daganzo, C. F.},
  journal={Transp. Res. Part B: Methodol.},
  volume={28},
  pages={269--287},
  year={1994}
}

@article{NEWELL1993281,
  title={A simplified theory of kinematic waves in highway traffic, part {I}: General theory},
  author={Newell, G. F.},
  journal={Transp. Res. Part B: Methodol.},
  volume={27},
  pages={281--287},
  year={1993}
}

@inproceedings{miao2025opt,
  title={{Opt-ODENet}: Neural {ODE} controller design with differentiable optimization layers for safety and stability},
  author={Miao, Keyan and Zhao, Liqun and Wang, Han and Gatsis, Konstantinos and Papachristodoulou, Antonis},
  booktitle={Proc. Learn. Dyn. Control Conf. (L4DC)},
  pages={1217--1229},
  year={2025}
}

@inproceedings{wilkman2025online,
  title={Online traffic density estimation using physics-informed neural networks},
  author={Wilkman, Dennis and Morozovska, Kateryna and Johansson, Karl Henrik and Barreau, Matthieu},
  booktitle={Proc. IEEE Conf. Decis. Control (CDC)},
  pages={236--242},
  year={2025}
}

@book{Goodfellow-et-al-2016,
  title={Deep Learning},
  author={Goodfellow, Ian and Bengio, Yoshua and Courville, Aaron},
  publisher={MIT Press},
  year={2016}
}

@book{ferrara2018freeway,
  title={Freeway traffic modelling and control},
  author={Ferrara, Antonella and Sacone, Simona and Siri, Silvia},
  publisher={Springer},
  year={2018}
}

@incollection{di2017follow,
  title={Follow-the-leader approximations of macroscopic models for vehicular and pedestrian flows},
  author={Di Francesco, Marco and Fagioli, Simone and Rosini, Massimiliano D and Russo, Giovanni},
  booktitle={Active Particles, Vol. 1},
  pages={333--378},
  publisher={Springer},
  year={2017}
}

@book{kerner2009introduction,
  title={Introduction to modern traffic flow theory and control},
  author={Kerner, Boris S},
  publisher={Springer},
  year={2009}
}

@article{delle2021three,
  title={A three-phase fundamental diagram from three-dimensional traffic data},
  author={Delle Monache, Maria Laura and others},
  journal={Axioms},
  volume={10},
  pages={17},
  year={2021}
}

@inproceedings{barreau2021physics,
  title={Physics-informed learning for identification and state reconstruction of traffic density},
  author={Barreau, Matthieu and Aguiar, Miguel and Liu, John and Johansson, Karl Henrik},
  booktitle={Proc. IEEE Conf. Decis. Control (CDC)},
  pages={2653--2658},
  year={2021}
}

@article{qu2017stochastic,
  title={On the stochastic fundamental diagram for freeway traffic: Model development, analytical properties, validation, and extensive applications},
  author={Qu, Xiaobo and Zhang, Jin and Wang, Shuaian},
  journal={Transp. Res. Part B: Methodol.},
  volume={104},
  pages={256--271},
  year={2017}
}

@article{treiber2000congested,
  title={Congested traffic states in empirical observations and microscopic simulations},
  author={Treiber, Martin and Hennecke, Ansgar and Helbing, Dirk},
  journal={Phys. Rev. E},
  volume={62},
  pages={1805},
  year={2000}
}

@article{rackauckas2020universal,
  title={Universal differential equations for scientific machine learning},
  author={Rackauckas, Christopher and others},
  journal={arXiv:2001.04385},
  year={2020}
}

@article{lagaris1998artificial,
  title={Artificial neural networks for solving ordinary and partial differential equations},
  author={Lagaris, Isaac E and Likas, Aristidis and Fotiadis, Dimitrios I},
  journal={IEEE Trans. Neural Netw.},
  volume={9},
  pages={987--1000},
  year={1998}
}

@article{chen2023hard,
  title={A hard-constraint wide-body physics-informed neural network model for solving multiple cases in forward problems for partial differential equations},
  author={Chen, Simin and Liu, Zhixiang and Zhang, Wenbo and Yang, Jinkun},
  journal={Appl. Sci.},
  volume={14},
  pages={189},
  year={2023}
}

@book{pontryagin2018mathematical,
  title={Mathematical theory of optimal processes},
  author={Pontryagin, Lev Semenovich},
  publisher={Routledge},
  year={2018}
}

@article{stern2018dissipation,
  title={Dissipation of stop-and-go waves via control of autonomous vehicles: Field experiments},
  author={Stern, Raphael E and others},
  journal={Transp. Res. Part C: Emerg. Technol.},
  volume={89},
  pages={205--221},
  year={2018}
}

@article{daniels2010monotone,
  title={Monotone and partially monotone neural networks},
  author={Daniels, Hennie and Velikova, Marina},
  journal={IEEE Trans. Neural Netw.},
  volume={21},
  pages={906--917},
  year={2010}
}

@article{lee2025traffic,
  title={Traffic control via connected and automated vehicles ({CAVs}): An open-road field experiment with 100 {CAVs}},
  author={Lee, Jonathan W and others},
  journal={IEEE Control Syst. Mag.},
  volume={45},
  pages={28--60},
  year={2025}
}

@article{shi2021physics,
  title={A physics-informed deep learning paradigm for traffic state and fundamental diagram estimation},
  author={Shi, Rongye and Mo, Zhaobin and Huang, Kuang and Di, Xuan and Du, Qiang},
  journal={IEEE Trans. Intell. Transp. Syst.},
  volume={23},
  pages={11688--11698},
  year={2021}
}

@article{di2023physics,
  title={Physics-informed deep learning for traffic state estimation: A survey and the outlook},
  author={Di, Xuan and Shi, Rongye and Mo, Zhaobin and Fu, Yongjie},
  journal={Algorithms},
  volume={16},
  pages={305},
  year={2023}
}

@inproceedings{liang2021modeling,
  title={Modeling trajectories with neural ordinary differential equations},
  author={Liang, Yuxuan and Ouyang, Kun and Yan, Hanshu and Wang, Yiwei and Tong, Zekun and Zimmermann, Roger},
  booktitle={Proc. IJCAI},
  pages={1498--1504},
  year={2021}
}

@article{ma2015long,
  title={Long short-term memory neural network for traffic speed prediction using remote microwave sensor data},
  author={Ma, Xiaolei and Tao, Zhimin and Wang, Yinhai and Yu, Haiyang and Wang, Yunpeng},
  journal={Transp. Res. Part C: Emerg. Technol.},
  volume={54},
  pages={187--197},
  year={2015}
}

@inproceedings{baloul2025traffic,
  title={Traffic flow reconstruction from limited collected data},
  author={Baloul, Nail and Hayat, Amaury and Liard, Thibault and Lissy, Pierre},
  booktitle={Proc. IEEE Conf. Decis. Control (CDC)},
  pages={7981--7986},
  year={2025}
}

@article{oliveira2012online,
  title={Online license plate matching procedures using license-plate recognition machines and new weighted edit distance},
  author={Oliveira-Neto, Francisco Moraes and Han, Lee D and Jeong, Myong K},
  journal={Transp. Res. Part C: Emerg. Technol.},
  volume={21},
  pages={306--320},
  year={2012}
}

@misc{securiteroutiere_radarsfixes,
  author       = {{S{\'e}curit{\'e} routi{\`e}re}},
  title        = {Radars de contr{\^o}le de la vitesse fixes},
  howpublished = {\url{https://www.securite-routiere.gouv.fr/radars/differents-types-de-radars/radars-fixes/radars-de-controle-de-la-vitesse-fixes}},
  note         = {Accessed: Nov. 2026}
}

@article{herrera2010evaluation,
  title={Evaluation of traffic data obtained via {GPS}-enabled mobile phones: The {Mobile Century} field experiment},
  author={Herrera, Juan C and Work, Daniel B and Herring, Ryan and Ban, Xuegang Jeff and Jacobson, Quinn and Bayen, Alexandre M},
  journal={Transp. Res. Part C: Emerg. Technol.},
  volume={18},
  pages={568--583},
  year={2010}
}

@article{hornik1991approximation,
  title={Approximation capabilities of multilayer feedforward networks},
  author={Hornik, Kurt},
  journal={Neural Netw.},
  volume={4},
  pages={251--257},
  year={1991}
}

@article{kingma2014adam,
  title={Adam: A method for stochastic optimization},
  author={Kingma, Diederik P and Ba, Jimmy},
  journal={arXiv:1412.6980},
  year={2014}
}
